\pdfoutput=1
\documentclass[11pt]{article}

\usepackage[final]{acl}
\usepackage{times}
\usepackage{latexsym}

\usepackage[T1]{fontenc}
\usepackage[utf8]{inputenc}

\usepackage{microtype}

\usepackage{inconsolata}

\usepackage{graphicx}

\usepackage{times,latexsym}
\usepackage{url}
\usepackage[T1]{fontenc}
\usepackage{amssymb}
\usepackage{color}
\usepackage{tabularray}
\usepackage{booktabs}
\usepackage{pifont}
\usepackage{graphicx}
\usepackage{hyperref}
\usepackage{amsmath}
\usepackage{multirow}
\usepackage{colortbl}
\usepackage{makecell} 

\definecolor{LightCyan}{rgb}{0.88,1,1}
\definecolor{LighterCyan}{rgb}{0.97,1,1}

\usepackage{geometry}
\usepackage{float}

\title{Survey on Question Answering over Visually Rich Documents: Methods, Challenges, and Trends}

\author{
  Camille Barboule
  \\
  Orange
  \\
  Paris, France
  \\
\And
    Benjamin Piwowarski
    \\
    Sorbonne Université, CNRS, ISIR
    \\
    Paris, France
    \\
\And
    Yoan Chabot
    \\
    Orange
    \\
    Belfort, France
    \\
}

\begin{document}
\maketitle
\begin{abstract}
The field of visually-rich document understanding, which involves interacting with visually-rich documents (whether scanned or born-digital), is rapidly evolving and still lacks consensus on several key aspects of the processing pipeline. In this work, we provide a comprehensive overview of state-of-the-art approaches, emphasizing their strengths and limitations, pointing out the main challenges in the field, and proposing promising research directions.
\end{abstract}

\section{Introduction}
Visually-rich documents (VRDs) combine complex information, blending text with visual elements like graphics, diagrams, and tables to convey detailed content effectively \cite{ding2024deeplearningbasedvisually}. Unlike traditional text documents, VRDs have two main features: text associated with typographic details (e.g., font, size, style, color), layout that organize information spatially, and visual elements, such as charts and figures, which enhance comprehension \cite{surveyfigures}. These documents can be either native digital files (e.g., PDFs) containing searchable text and layout metadata, or scanned images requiring OCR to extract text and layout. Visually-rich Document Understanding (VrDU) is a rapidly evolving field at the intersection of computer vision and natural language processing, tackling both perception (document parsing, i.e. identification and extraction of objects within the document) and interpretation (downstream tasks using the document features, such as answering questions or information extraction) \cite{zhang2024documentparsingunveiledtechniques}.

% Question-answering tasks on VrDU can be simply extractive, or requiring reasoning skills such as arithmetic, logical, and spatial reasoning, as well as counting, sorting, comparison, and multi-hop reasoning, which may involve combining multiple elements to answer a question, potentially spanning across modalities, pages, or even documents, as depicted in Table~\ref{tab:datasets}.

We provide a comprehensive analysis of how Visual Document Understanding (VrDU) models represent visually rich documents (VrDs) and use these features on downstream tasks, which often contain multiple elements—such as charts, tables, figures, and text—and span multiple pages (see Table~\ref{tab:datasets} in appendix). Current VrDU approaches typically follow a two-step pipeline: document parsing followed by downstream tasks like question answering. %This process can be either explicit, where the output of a document parsing model (e.g., OCR or metadata from PDFs) is passed to a language model, or implicit, where a model is pretrained on a document parsing task and later fine-tuned for question answering. 
We analyze how this two-step pipeline operates, looking first at how VrDU models encode VrDs, and then how large language models (LLMs) decode those features for downstream tasks.

% We first take a deep dive into current approaches for processing and leveraging tokens and bounding boxes (extracted from OCR or PDF metadata) and linking textual and visual features within documents. Recent innovations aim to address these challenges, enabling large language models (LLMs) to better understand the structure and content of VrDs (Section~\ref{sec:t+l+v}).

We first take a deep dive into current approaches for processing and leveraging tokens and bounding boxes (extracted from OCR or PDF metadata) and linking textual and visual features within documents. Recent innovations aim to enable LLMs to handle the 2D positioning of elements in VrDs at different granularities and to process both textual and visual features from those documents, thereby improving their understanding of the structure and content of VrDs (Section~\ref{sec:t+l+v}).

Additionally, we examine how Large Vision-Language Models (LVLMs), which are increasingly recognized for their combined perception and reasoning capabilities, currently dominate the VrDU domain. Recent innovations focus on balancing coarse- and fine-grained visual representations of VrDs while limiting computational cost. Despite their growing popularity, we show that current LVLM architectures are still ill-suited to the specific challenges of VrDU, particularly in handling multi-page documents (Section~\ref{sec:v-only}).

Next, we analyze how VrDU approaches handle multi-page documents, exploring recent page-by-page strategies, strategies relying on sparse attention mechanisms to maintain connections across pages, and we finally examine retrieval-augmented generation (RAG) approaches that reduce the problem to a single-page context by retrieving relevant information from other pages, while giving insights on future promising directions (Section~\ref{sec:multipage}).

Finally, we compare the different approaches to optimally inject those visual information into a LLM to be processed optimally for downstream tasks, comparing self-attention and cross-attention-based approaches (Section~\ref{sec:decoding}).

\section{Encoding VrDs from structured information}
\label{sec:t+l+v}
VrDs can be represented through three distinct but interconnected features: text and layout, derived from native digital formats or OCR extraction, and the overall visual appearance of the document, obtained by generating a screenshot of the document page. The most important layout features are bounding boxes around text and structural elements (e.g., tables). The visual modality captures the document page appearance, encompassing the overall structure and visual context of the document as a whole. The main problematic in VRD encoding is to represent and merge the information coming from these three distinct modalities. Table~\ref{tab:t+l+v} summarizes models from this category that we detail in this section.
\begin{table*}[!t]
\centering
\resizebox{\textwidth}{!}{
   \begin{tabular}{|lcccccc|}
    \toprule
\textbf{Model} & $\mathbf{E_{Text}}$ & $\mathbf{E_{Vis}}$ & $\mathbf{E_{Pos}}$ & $\mathbf{E_{Cross}}$ & $\mathbf{D_{Text}}$ & \textbf{MP} \\

\midrule
\rowcolor{LightCyan} \multicolumn{7}{c}{\textbf{Interaction of text and visual features within self-attention after modalities concatenation}} \\ 
\midrule

LayoutLMv2 \citeyear{layoutlmv2} & UniLMv2 & ResNeXt-101-FPN & emb. tables + attn bias & & transformer & \\
LayoutXLM \citeyear{layoutxlm} & XLM-R & ResNeXt-101-FPN & emb. tables + attn bias & & transformer & \\
UNITER \citeyear{uniter} & BERT & Faster R-CNN & emb. tables (7D) & & transformer & \\
LayoutLMv3 \citeyear{layoutlmv3} & RoBERTa & ViT & attn bias & & transformer & \\
DocFormerv2 \citeyear{docformerv2} & T5 encoder & ViT & emb. tables. & & T5 & \\
GRAM \citeyear{gram} & DocFormerv2(\citeyear{docformerv2}) & DocFormerv2(\citeyear{docformerv2}) & emb. tables & & DocFormerv2(\citeyear{docformerv2}) & \checkmark \\
LayoutLLM \citeyear{layoutllm} & LayoutLMv3(\citeyear{layoutlmv3}) & LayoutLMv3(\citeyear{layoutlmv3}) & LayoutLMv3(\citeyear{layoutlmv3}) & & Llama-7B & \\
DocLayLLM \citeyear{doclayllm} & LayoutLMv3(\citeyear{layoutlmv3}) & LayoutLMv3(\citeyear{layoutlmv3}) & LayoutLMv3(\citeyear{layoutlmv3}) & & Llama3-8BInstruct & \\

\midrule
\rowcolor{LightCyan} \multicolumn{7}{c}{\textbf{Interaction of text and visual features within cross-attention}} \\
\midrule

DocFormer \citeyear{docformer2021} & LayoutLM(\citeyear{layoutlm}) & ResNet50 & emb. tables & visual-spatial attn & transformer & \\
SelfDoc \citeyear{selfdoc} & Sentence BERT & Faster R-CNN & emb. tables & intra\&inter-modal attn & transformer & \\
ERNIE-Layout \citeyear{ernie} & BERT & Faster R-CNN & emb. tables & Disentangled attn (\citeyear{he2021debertadecodingenhancedbertdisentangled}) & transformer & \\
HiVT5 \citeyear{hivt5} & T5 encoder & DiT (\citeyear{dit}) & emb. tables & VT5 encoder & VT5 decoder & \checkmark \\
DocTr \citeyear{doctr} & LayoutLM(\citeyear{layoutlm}) & DETR (\citeyear{detr}) & special tokens & Deformable DETR (\citeyear{zhu2021deformabledetrdeformabletransformers}) & LayoutLM  & \\
InstructDr \citeyear{instructdoc} & FlanT5 encoder & CLIP VIT-L/14 & emb. tables & Document-Former & FlanT5 & \checkmark \\
RM-T5 \citeyear{Dong2024MultipageDV} & T5 encoder & DiT (\citeyear{dit}) & emb. tables & RMT (\citeyear{recurrentmemory}) & T5 decoder & \checkmark \\
Arctic-TILT \citeyear{arctictilt} & T5 encoder & U-Net (per RoI) & attn bias & Tensor Product & T5 & \checkmark \\

\midrule
\rowcolor{LightCyan} \multicolumn{7}{c}{\textbf{Summing aligned text and visual features via ROI-pooling}} \\ 
\midrule

TILT \citeyear{tilt2021} & T5 encoder & U-Net & attn bias & & T5 & \\
\citet{pramanik} & Longformer & ResNet50 + FPN & sinusoidal emb. & & transformer & \checkmark \\
UDOP \citeyear{udop} & T5 encoder & MAE encoder & attn bias & & T5\&MAE decoder & \\

% \midrule
% \rowcolor{LightCyan} \multicolumn{7}{c}{\textbf{Interaction of text and visual features with hierarchical / graph architectures}} \\ 
% \midrule

% PICK \citeyear{pick} & transformer & CNN & & GNN & BiLSTM + CRF & \\
% GraphDoc \citeyear{Zhang2022MultimodalPB} & SentenceBERT & SwinT + FPN & emb. tables & GAT (\citeyear{gats}) & transformer & \\
% LAMPreT \citeyear{lampret} & BERT & CNN &  & Hierarchical fusion & transformer & \\

\bottomrule
\end{tabular}
}
\caption{Comparison of VrDU models handling the three modalities (T+L+V), detailing encoding of text $\mathbf{E_{Text}}$, visuals $\mathbf{E_{Vis}}$, and position $\mathbf{E_{Pos}}$, fusion layers $\mathbf{E_{Cross}}$, decoder $\mathbf{D_{Text}}$, and multi-page (MP) support \ding{51}.}
\label{tab:t+l+v}
\end{table*}

\subsection{Integrating the Layout information}

The positions and sizes of elements within a document can vary in granularity, from individual tokens \cite{lambert2021, layoutlm} to larger blocks like cells, tables, images, or paragraphs \cite{structurallm, selfdoc}. This layout information can be represented within VrDU models in three ways: through absolute positional embeddings of the 2D position, as an attention bias / rotation depending on the spatial distance of the tokens, or directly within the text, as special tokens.

The simplest approach, which does not require any architectural change, is to include layout information as special tokens,  directly within the text \cite{lu2024lyricsboostingfinegrainedlanguagevision, vitlp}. The global text-layout sequence is based on an extended vocabulary \(\hat{V} = V \cup [\text{BBOX}]\), where $V$ is the original text vocabulary. This approach not only increases the sequence length, overloading the model's context window, but also limits the ability to capture complex spatial interactions between elements in the document.

This is why the VrDU community has focused on developing optimal methods to incorporate spatial information of tokens within documents. One way is to extend the 1D absolute positional encoding of tokens in transformers to 2D (see Table~\ref{tab:t+l+v}) by embedding the spatial coordinates $(x, y)$ of each token's bounding box. For example, LayoutLM \cite{layoutlm} embeds the discretized $x$ and $y$ coordinates separately and sums them. DocFormer \cite{docformer2021} further includes embeddings for the bounding box dimensions (height and width), while UNITER \cite{uniter} adds an embedding for the area of each bounding box. These embeddings can be learned or fixed (function-based, e.g., sinusoidal \cite{hong2022brospretrainedlanguagemodel}). 

However, absolute positional encoding is limited, as they are added at the input only \cite{chen2021simpleeffectivepositionalencoding}. Recent models hence apply positional encoding directly within the attention mechanism for improved performance and flexibility.
In particular, they extend the relative positional encoding \cite{press2022trainshorttestlong, raffel2023exploringlimitstransferlearning}, applied on every self-attention layers, to a 2D space. Such approaches either encode the 2D distance as a bias term added before the softmax, representing the horizontal and vertical distances between tokens within the document \cite{layoutlmv2, tilt2021}, or as a rotation applied to the queries and keys vectors, depending on the absolute position of each token, inspired from 1D-RoPE \cite{su2023roformerenhancedtransformerrotary}, with a rotation of the attention score depending on the horizontal position of the token (e.g. position within a table row), and another on the vertical one (e.g. position within the columns of the table), with both scores weighted by a gating model \cite{li20242dtpetwodimensionalpositionalencoding}. Pondering the attention score with the 2D distance of the tokens is still limited, as token semantics, like "total" in tables, often dictate specific spatial interactions beyond mere positional proximity. To ensure that the model pays particular attention to tokens located at the same horizontal position of some meaningful tokens (like "total" in a table), ERNIE-Layout \cite{ernie} introduces three relative position attention biases (disentangled attention), capturing respectively how the semantic meaning of a token interacts with its sequential, horizontal and vertical relative distance to the other token. FormNet \cite{formnet} goes further in this direction by allowing more complex interactions, using functions that combine semantic and position information between tokens.

To conclude, in a world where documents are increasingly digital-native, with direct access to text and bounding boxes, enabling LLMs to handle such structures is crucial. However, the community has mostly focused on adapting either 1D absolute positional encodings or relative 1D positional bias to the 2D space, while little attention has been given to extending RoPE to 2D—despite most current models relying on it.

% The performance of these multimodal models (text, layout, visual) suggests that a multi-granular approach is necessary for document modeling. Specifically, regions representing semantically meaningful components or units within a document should be modeled based on their 2D positional relationships. Simultaneously, the token-level should also be taken into account, considering the sequential relationships between tokens within each component \cite{wang2022mgdocpretrainingmultigranularhierarchy}.
To the best of our knowledge, only a few studies focus on the granularity of positional information, distinguishing between intra-region positions (e.g., the position of a cell within a table or a token within a paragraph) and page-level positions (e.g., the position of a token or a region within the entire page).
Region-level models fail to capture cross-region and word-level interactions, while page-level models (with token-wise positions) suffer from excessive contextualization \cite{selfdoc}. We suggest that combining these two levels of granularity could enhance performance \cite{wang2022mgdocpretrainingmultigranularhierarchy}.

\subsection{Integrating the visual information}

In all the works we reviewed, the visual modality is transmitted as a set of visual ``tokens'' (vectors), computed by a visual encoder. Initially based on CNNs \cite{layoutlmv2}, these encoders have transitioned to Visual Transformers (ViTs) \cite{layoutlmv3}. 

Fusing text and visual features for unified document encoding is challenging due to the differences between visual and text tokens (see Table~\ref{tab:t+l+v}). The integration of the two modalities can be done locally (per regions or the document) or globally (within the whole document).

Global modality alignment involves considering both the visual and textual features of the entire document rather than specific regions. A simple method to align those modalities globally involves concatenating them~\cite{layoutlmv2}. A transformer encoder then allows interaction through standard self-attention mechanisms \cite{docformerv2, layoutlmv3}. However, such approaches require intensive pretraining for features (visual and textual) alignment \cite{layoutlmv3}, since these two feature types form a unit within the document, sometimes representing the same elements (e.g., an image of a piece of text versus the text itself).

Local modality alignment refers to aligning text and visual features specifically within localized regions of the document, focusing solely on the text and visual attributes from those regions. These regions can be either inferred using  visual information, i.e. determined by an object detection module \cite{detr, fasterrcnn} or determined by the textual information, i.e. considering the bounding boxes of text tokens \cite{tilt2021}. A simple method to locally align modalities involves summing the two representations per region \cite{tilt2021}. Note that regions without associated text only have a vision-only representation \cite{udop}. However, this approach constrains the interaction between visual and textual modalities, thereby limiting the comprehensive understanding of the document region \cite{selfdoc}.

To capture interactions between textual and visual features from a region of the document, SelfDoc \cite{selfdoc} uses two cross-attentions: from the visual to textual tokens and vice-versa, e.g. allowing the textual semantic representation to be contextualized by visual information such as color, bold elements, and position. For example, a large, bolded, centered text block is likely to serve as a title or header. By incorporating these visual cues, the model refines the semantic representation of text, ensuring that its meaning is informed by its visual context within the document. Rather than relying on costly cross-attention for modality fusion and interaction, Arctic-TILT \cite{arctictilt} introduces a lightweight attention mechanism after the transformer feed-forward layer to integrate visual information using a learnable role bias for text tokens, inspired by TP-Attention \cite{schlag2020enhancingtransformerexplicitrelational}.

% Finally, visual and textual tokens can interact by viewing the document as a graph, built thanks to the bounding boxes of elements like paragraphs, tables or figures, each representing a node. Graph-Based Relationship Modeling employs graph neural networks (GNNs) to capture relationships between modalities in documents \cite{pick}. Various graph-based approaches differ in how they aggregate neighbor information between nodes: Techniques include convolution (Graph Convolutional Networks) \cite{pick}, attention (Graph Attention Networks) \cite{Li2023EnhancingVD}, and other aggregation methods \cite{formnet}. While graph-based modeling is particularly effective at capturing long-range dependencies and contextual relationships in documents, it remains limited in facilitating interactions between the semantic of modalities, since modalities interaction is only layout-based.

% Furthermore, efforts must be done to align the textual and visual modalities (many pretraining strategies aim at aligning tokens with visual features from the document, like Text-Image Alignment \cite{layoutlmv2}, Word-Patch Alignment \cite{layoutlmv3}, ...).

To conclude, the effect of the visual features, at least in the way it is utilized in such models (i.e. enriching the textual features' representation), appears small and may primarily introduce redundancy to the textual elements: as shown by \citet{udop}, adding visual features brings little to no improvement on datasets without images or visual components, and only marginally enhances performance on highly visual tasks like InfographicsVQA \cite{infographicvqa}.

\section{Vision-Only Encoding of VrDs}
\label{sec:v-only}

\begin{table*}[!t]
\centering
\resizebox{\textwidth}{!}{
   \begin{tabular}{|l|ccccc|}
    \toprule
    \textbf{Model} & \textbf{Res.} & $\mathbf{E_{Vis}}$ & $\mathbf{P_{E_V\to D_T}}$ & $\mathbf{D_{Text}}$ & \textbf{MP} \\
    \midrule
    
    \rowcolor{LighterCyan} \multicolumn{6}{c}{\textbf{Encoder: HR image -- Decoder: Tiny Decoder}} \\ 
    \midrule
    DONUT \cite{donut} & 2560x1920 & SwinT (\citeyear{swinvit}) & MLP & BART & \\
    DESSURT \cite{dessurt} & 1152x768 & Attn-Based CNN & MLP & BART with Swin attn & \\
    Pix2Struct \cite{pix2struct} & 1024x1024 & ViT & MLP & BART & \\
    SeRum \cite{cao2023attentionmattersrethinkingvisual} & 1280x960 & SwinT \citeyear{swinvit} & MLP & mBART & \\
    Kosmos2.5 \cite{kosmos25Paper} & 224x224 & Pix2Struct \citeyear{pix2struct}'s ViT & Perceiver Resampler & Transformer & \\
    
    \midrule
    \rowcolor{LighterCyan} \multicolumn{6}{c}{\textbf{Encoder: LR image -- Decoder: LLM}} \\ 
    \midrule
    LLaVAR \cite{llavar} & 336x336 & CLIP VIT-L/14 & MLP & Vicuna13B & \\
    Unidoc \cite{unidoc} & 336x336 & CLIP VIT-L/14 & MLP & Vicuna13B & \\
    mPLUG-DocOwl \cite{ye2023mplugdocowlmodularizedmultimodallarge} & 224x224 & CLIP VIT-L/14 & Visual Abstractor & Llama-7b & \\
    QwenVL \cite{bai2023qwenvlversatilevisionlanguagemodel} & 448x448 & CLIP-VIT-G/14 & Cross-attn layer & Qwen-7b & \\
    
    \midrule
    \rowcolor{LightCyan} \multicolumn{6}{c}{\textbf{Encoder: HR image -- Decoder: LLM thanks to HR image in subimages division (Section~\ref{croping-methods})}} \\ 
    \midrule
    \multirow{2}{*}{SPHINX \cite{lin2023sphinxjointmixingweights}} & \multirow{2}{*}{1344x896} & VIT \& ConvNext \& & \multirow{2}{*}{MLP} & \multirow{2}{*}{Llama2-7B} & \\
    & & DINO \& QFormer & & & \\
    UREADER \cite{ureader} & 2240x1792 & CLIP ViT-L/14 & MLP & Vicuna13B & \\
    Monkey \cite{monkey} & 1344x896 & CLIP Vit-BigG & Perceiver Resampler & Qwen-7B & \\
    TextMonkey \cite{textmonkey} & 1344x896 & CLIP Vit-BigG & Shared Perceiver Resampler & Qwen-7B & \\
    mPLUG-DocOwl1.5 \cite{mplugdocowl1.5} & 2560x1920 & EVA-CLIP & H-Reducer & Llama-7b + MAM & \\
    LLaVA-UHD \cite{xu2024llavauhdlmmperceivingaspect} & 672x1088 & CLIP-ViT-L & Shared perceiver Resampler & Vicuna-13B & \\
    InternLMXC2-4KHD \cite{internlmxcomposer24khd} & 3840x1600 & CLIP-ViT-L & PLoRA matrix & InternLM2-7B & \\
    Idefics2 \cite{laurençon2024mattersbuildingvisionlanguagemodels} & 980x980 & SigLIP-SO400M & MLP & Mistral-7B-v0.1 & \\
    TextHawk \cite{texthawk} & 1344x1344 & SigLIP-SO & Perceiver Resampler & InternLM-7B & \\
    TokenPacker \cite{tokenpacker} & 1344x1344 & CLIP-ViT-L & TokenPacker & Vicuna-13B & \\
    mPLUG-DocOwl2 \cite{mplugdocowl2} & 504x504 & EVA-CLIP & H-Reducer+DocCompressor & Llama-7b + MAM & \checkmark \\
    
    \midrule
    \rowcolor{LightCyan} \multicolumn{6}{c}{\textbf{Encoder: HR image -- Decoder: LLM thanks to adaptation of ViT to capture fine-grained details (Section~\ref{sec:modify-vit})}} \\ 
    \midrule
    DocPedia \cite{feng2024docpediaunleashingpowerlarge} & 2560x2560 & SwinT \citeyear{swinvitv2} & MLP & Vicuna-13B & \\
    LLaVA-PruMerge \cite{llava-prumerge} & 336x336 & CLIP-ViT & MLP & Vicuna13B & \\
    CogAgent \cite{hong2024cogagentvisuallanguagemodel} & 1120x1120 & EVA2-CLIP \& CogVLM & Cross-attn layer \& MLP & Vicuna-13B & \\
    Vary \cite{vary} & 1024x1024 & ViTDet \& CLIP-ViT-L & MLP & Qwen-7B & \\
    Mini-Gemini \cite{mini-gemini} & 2048x2048 & ConvNeXt \& ViT-L/14 & MLP & Mistral-7B & \\
    LLaVA-HR \cite{luo2024feasteyesmixtureofresolutionadaptation} & 1024x1024 & CLIP-ConvNeXt \& ViT-L & MLP \& MR-Adapter & Llama2-7B & \\
    TinyChart \cite{tinychart} & 768x768 & SigLIP & MLP & Phi-2 & \\
    HRVDA \cite{hrvda} & 1536x1536 & SwinT (\citeyear{swinvitv2}) & MLP & Llama2-7B & \\
    DocKylin \cite{dockylin} & 1728x1728 & SwinT (\citeyear{swinvitv2}) & MLP & Qwen-7B & \\
    \bottomrule
    \end{tabular}
    }
\caption{Comparison of vision-only VrDU models, detailing the input image resolution (Res), visual encoding $\mathbf{E_{Vis}}$, vision-to-text projection $\mathbf{P_{E_V\to D_T}}$, decoder $\mathbf{D_{Text}}$, and multi-page (MP) support \ding{51}.}
\label{tab:v-only}
\end{table*}

In the previous section, we discussed techniques that integrate visual and textual information. These models however remain complex because the segmentation between modalities in a document is not straightforward and may introduce redundancy, lead to information loss and require pretraining for modalities alignment.

Many recent works consider VrDs as images, which brings the advantage of dealing with a single modality, relying on a LLM decoder to handle different tasks. A summary of this type of model we detail below is provided in Table~\ref{tab:v-only}. 

Such approaches, commonly named Large Visual-Language Models (LVLMs), demand a highly capable visual encoder to capture all textual, layout, and visual details within the document. However, ViTs themselves are not capable to capture fine details like text \cite{zhang2025mllms}. Indeed, in ViTs, the visual input (e.g., a document page) is divided into fixed-size patches, each becoming a "vision token" (e.g., 14x14 or 16x16 pixels). If patches are too large, they may cover too much content, like multiple lines or text fragments, and miss fine details. Using smaller patches or increasing the image resolution creates more patches, enabling the model to capture finer details and better encode the document's textual content \cite{pix2struct}, but at the cost of efficiency.

Indeed, ViTs have a maximum context size (number of patches) they can manage \cite{pix2struct}. This is why research in vision-only VrDU focuses on architectural modifications to ViTs to enable the processing of high-resolution images (Section~\ref{sec:modify-vit}). An effective alternative is to use a set of pre-trained ViTs, each handling a different part of the image, thereby allowing the processing of high-resolution images more efficiently (Section~\ref{croping-methods}). In this case, it is necessary to ensure coherence between the cropped regions of the page.

\subsection{Architectural changes to ViT}
\label{sec:modify-vit}

A number of approaches leverage CNN architectures, which capture local information more efficiently than ViTs due to their intrinsic design based on convolutions, exploiting locality bias in images. \citet{dhouib2023docparserendtoendocrfreeinformation} proposes a sequential architecture combining CNN and ViT components, where ConvNext blocks are used to extract local features, and their output is fed into a ViT for modeling global dependencies.

Due to the complexity of combining two networks without losing information, other approaches \cite{donut, nougat} draw inspiration from the local window mechanism of CNNs and incorporate it into ViTs, enabling them to process numerous patches effectively. These approaches restrict attention to a local window of patches with a Swin Transformers \cite{swinvit}, which applies self-attention within local windows, shifting these windows across layers to efficiently integrate cross-window information.  However, Swin ViTs progressively reduce the resolution of the tokens through token merging steps, which decrease the number of tokens. DocPedia \cite{feng2024docpediaunleashingpowerlarge} removes this  downsampling step, keeping the full token resolution throughout the processing pipeline by leveraging the frequency domain rather than spatially merging patches as done in Swin. More precisely, they represent an image in the frequency domain, using the Discrete Cosine Transform \cite{frequencyvit}, allowing to process larger patches without loosing important high resolution information. However, restricting the attention to local windows, even if shifted, introduces a locality bias to ViTs, similar to CNNs.

More recent approaches avoid introducing a locality bias to ViTs, instead focusing on removing redundant information from ViT patches, as documents often contain a significant amount of redundancies, such as borders, whitespace or decorations. These methods either use attention scores from the self-attention mechanism to prune or merge tokens (e.g., \citet{tinychart, llava-prumerge, chen2024imageworth12tokens}) or employ unsupervised techniques like Dual-Center K-Means Clustering \cite{dockylin} to select tokens. % \citet{llava-prumerge} perform token selection only in the final layer, using the attention scores between the class token and vision tokens, while \citet{chen2024imageworth12tokens} discard tokens in deeper layers, focusing attention on "anchor" tokens. 
TinyChart \cite{tinychart} combines similar tokens after each ViT layer using methods like average pooling, while DocKylin \cite{dockylin} employs similarity-weighted summation based on token cosine similarity ensuring that each token contributes proportionally to its relevance. Other approaches \cite{hrvda} use a content detection module to filter out low-relevance areas (e.g., whitespace) and preserve meaningful regions (e.g., text or tables) by assigning probabilities to pixels and mapping them to patches.

% DocKylin \cite{dockylin} goes further in the pixel-level direction of HRVDA \cite{hrvda}, before the image tokenization, to identify redundant information: A preprocessing module (Adaptive Pixel Slimming) modifies the image itself, removing whitespace by detecting low-gradient areas (smooth regions lacking significant content). The identified redundant rows and columns are then cropped, resulting in a more compact and efficient document representation.

\subsection{Several ViTs to process partitioned image}
\label{croping-methods}

Recent works have explored pipelines leveraging already pretrained ViTs to process high-resolution images cut into slices. Each ViT handles a specific portion of the image, and the resulting representations are combined (sequence of "image tokens") as the unified document representation.

The way the original image is sliced into subimages is crucial to prevent information loss. Padding preserves the aspect ratio and prevents deformation \cite{tokenpacker}. Some approaches predict the optimal way to cut the original image, with pre-defined grid matching \cite{ureader} and a score function predicting the best partition \cite{llava-uhd}, resulting in a varying amount of crop. Whatever the method, models need to maintain the continuity between the different subimages representations.

A simple way to do so is through a 2D crop position encoding, which allows interaction between local images~\cite{ureader}. However, this approach lacks information continuity between cropped images. To alleviate salient information loss due to cropping, \citet{textmonkey} introduces a Shifted Window Attention mechanism, enabling sliding window-based attention across subimage representations. %This shared representation layer takes into account the position of the crops within the image, either via scalable positional embeddings to handle varying subimages amount \cite{texthawk} or with learned embeddings \cite{ureader}. If different types of visual encoders are applied for each subimages \cite{lin2023sphinxjointmixingweights}, aligning those embeddings for a unified document representation is complex.

%Another approach to maintain continuity across subimages is to add a low-resolution global view of the entire document to retain overall context, which is encoded with an independent ViT, alongside the different subimages \cite{monkey}. However, treating the global view the same way as the subimages is not optimal since the two convey different types of information \cite{tokenpacker}. 

A more efficient approach to maintain continuity between subimages is to leverage a low-resolution document representation to guide the integration of subimages. Through a cross-attention layer, TokenPacker \cite{tokenpacker}, and later mPLUGDocOwl2 \cite{mplugdocowl2}, integrate the high-resolution representation of regions into the low-resolution representations using cross-attention, thus interpolating these low-resolution representations with its multi-level region cues treated as reference keys and values to inject their finer information to global image view.

%Dividing the image into sub-images indeed reduces training costs, but it is highly challenging to avoid information loss caused by the slicing process. Instead of "zooming" across the entire document and dividing it into patches, a more effective approach could involve an adaptive "zoom" focused on the areas where relevant information is located based on the query \cite{anonymous2024mllms}.

To conclude on vision-only approaches, we think that slicing approaches using local information from cropped image regions to complement a low-resolution global view are promising, enabling compact and efficient representations with significantly fewer tokens while maintaining essential layout and semantic details \cite{mplugdocowl2}. However, while this type of approach reduces computational cost for single-page processing, it is not sufficient to handle multi-page \cite{mplugdocowl2}.
% We also think that techniques like adaptive pixel slimming \cite{dockylin} are promising for VrDU, leveraging the inherent structure of documents, containing whitespace or decorations, to preserve all its content. 

\section{Encoding multi-pages documents}
\label{sec:multipage}
The principal challenge in VrDU is to handle multi-pages documents. Multi-page documents vary in length (e.g., 20 pages in SlideVQA \cite{slidevqa}), amount of tokens per document (e.g., 21214 tokens per document in MMLongBench-Doc \cite{mmlongbenchdoc}), and cross-page information, i.e. questions requiring information from several pages of the document (e.g. 2.1\% in DUDE \cite{dude}).
To encode multi-page documents, recent approaches use retrieval-augmented generation (RAG) techniques \cite{rag} (Section~\ref{sec:rag}). Other methods represent the document page by page (Section~\ref{pagetoken}), enhanced with inter-page interactions inherited from long-sequence processing techniques (Section~\ref{sec:longsequences}).

\subsection{Retrieval Approach to multi-page}
\label{sec:rag}
The retrieval approach to multi-page documents focuses on supplying to the VrDU decoder only the representation of pages with relevant information. Several levels can be used to identify the relevant element from the document: the retriever can either predict the entire relevant page \cite{multipageretrieval, faysse2024colpaliefficientdocumentretrieval, ma2024unifyingmultimodalretrievaldocument, cho2024m3docragmultimodalretrievalneed} or focus on specific regions within the page, such as paragraphs or images containing the elements to answer the question \cite{pdfwukong}.
%PDF-Wukong \cite{pdfwukong} employs a region-level approach, dividing the document into modalities such as text paragraphs and images for tables and charts. It trains a retriever (Sparse-Sampler) using contrastive learning to align the representations of "oracle" elements (text or images) with the query. The model uses the encoder from InternLMXComposer2\_4KHD \cite{internlmxcomposer24khd} for images and modifies the weights of the text encoder from BGE-M3 \cite{bgem3} to align with the visual representations. During inference, the representations of the positive elements and the query are passed to the decoder of InternLMXComposer2\_4KHD, which generates the answer. This region-level approach is limited because it encodes each modality separately, preventing any connection between them.
%In contrast, \cite{multipageretrieval} employs a page-level approach by encoding entire pages of the document with Pix2Struct-Base \cite{pix2struct}, using a self-attention scoring module to identify the most relevant page based on the query. This page-level approach is limited as it encodes each page independently, which prevents connections between pages.

These approaches inherently limit either the interaction between pages or the interaction between modalities, which does not allow cross-page analysis \cite{mmlongbenchdoc}, not mentioning that they highly depend on the performance of the retriever.
% Other possibilities, described in the next sections, can be used for multi-modal encoders, but they are however the only viable option to enable vision-only models to handle multi-page inputs (see Table~\ref{tab:vqa-performance}), since vision-only models encode a page with many tokens.

\subsection{Query-based approaches}
\label{pagetoken}

HiVT5 \cite{hivt5}, and later InstructDr \cite{instructdoc}, encode each page of the document separately, with a specific learnable token added at the start of each page. HiVT5 \cite{hivt5} uses the specialized \texttt{[PAGE]} tokens to guide the encoder in summarizing each document page based on the given question, by processing separately each page with the question, encoding all the relevant information for the next processing step into the \texttt{[PAGE]} token. These \texttt{[PAGE]} tokens representations are then concatenated and passed to the decoder to generate the final answer. 
To our knowledge, the only vision-only model designed for multi-page input is mPLUG-DocOwl2 \cite{mplugdocowl2}, which compresses each page representation into 324 tokens and adds a page token for each page. In vision-only approaches, the token length of high-resolution images (i.e., document pages) is typically too large for LLMs to handle multi-page joint understanding, necessitating extreme compression of each page representation and thus degrading performance \cite{mplugdocowl2}.

However, query-based approaches only allow limited cross-page reasoning, as the long sequence and diluted information across pages make it challenging to capture specific inter-page relationships \cite{mmlongbenchdoc}, the page token being not leveraged effectively.

\subsection{Efficient encoding of multi-pages}
\label{sec:longsequences}

Inspired by the ETC Global-Local Attention mechanism \cite{etc}, GRAM \cite{gram} enables global reasoning across multiple pages through a combination of page-dedicated layers, which apply self-attention within each page representation, and document-level layers, which focus exclusively on page token embeddings in their attention computations.

Another sparse attention approach is implemented by Arctic-Tilt \cite{arctictilt}, employing a blockwise attention strategy limiting the attention to a chunk size, allowing to handle up to 500 pages (about 390k tokens, with 780 tokens per page on average). This method limits attention to a smaller, predefined neighborhood ($\approx$2 pages), reducing complexity from quadratic to linear while representing cross-page information.

An alternative to sparse attention for efficient multi-page documents processing is to use a recurrent network. RM-T5 \cite{Dong2024MultipageDV} uses a Recurrent Memory Transformer (RMT) \cite{recurrentmemory} to process multi-page documents sequentially, treating each page as part of a sequence. This allows the model to carry information across pages by utilizing hidden states from previous pages. The RMT selectively retains or forgets information, capturing essential details from each page for the next encoder, with all memory cells concatenated for the decoder to generate the final answer. However, the drawbacks of RNNs are inherited, such as the lack of parallelization and the limited possible interaction of two elements (here, pages) distant in the sequence.

%DocMamba \cite{docmamba} replaces transformer with state space models (SSMs) \cite{mamba} to reduce computational complexity for multi-page document encoding while preserving global modeling capabilities.

Overall, our view is that approaches that encode entire documents using sparse attention techniques, either global-local or blockwise, represent the future of the multi-page field, as they show great performance on cross-page reasoning \cite{mmlongbenchdoc} over retrieval ones.

\section{Injecting visual features into the LLM}
\label{sec:decoding}

In both approaches for encoding the VrD (structured encoding in Section~\ref{sec:t+l+v} versus vision-only encoding in Section~\ref{sec:v-only}), the representation of the document contains visual features. Integrating visual features into an LLM decoder is not straightforward because it requires adapting the visual representation space into an LLM-compatible representation without losing information, while preserving some computational efficiency. We detail here how this integration is done by current VrDU approaches, and what the future directions for visual features integration into LLMs are.

\subsection{Self-attention based approach}
\label{selfattn}

This self-attention approach \cite{laurençon2024mattersbuildingvisionlanguagemodels} consists in  prepending the visual representation to the prompt, allowing the model to process both visual features with the prompt together in its self-attention layers. In such approaches, visual features are projected into the LLM space via several approaches, and are optionally pooled into a shorter sequence.

Those methods vary in complexity, ranging from direct linear projection using a single layer to map visual tokens to the expected input format of the language model \cite{pix2struct}, which minimizes the number of parameters; convolutional approaches \cite{honeybee}, which reduce the dimensionality of the visual representation; to using learnable queries \cite{blip2, bai2023qwenvlversatilevisionlanguagemodel}, used to retrieve relevant visual tokens.

%Another approach is to use cross-attention to integrate the VrD encoder output. In \cite{bai2023qwenvlversatilevisionlanguagemodel}, this cross-attention is called Visual Abstractor \cite{ye2023mplugdocowlmodularizedmultimodallarge}. The Q-Former module \cite{blip2} uses two submodules for visual features and the prompt, enabling interaction through self- and cross-attention.

Since interactions within visual tokens are already handled by the vision encoder in vision-only approaches, \citet{ee-mllm} modify the self-attention mechanism of the LLM by a Composite-Attention, removing interactions within the LLM within visual tokens; text tokens act as queries, with both visual and text tokens serving as keys and values.

These approaches are limited, considering raw tokens of the textual prompt and visual tokens from the document at the same level, without distinguishing between their respective roles or significance.

\subsection{Cross-attention based approach}
\label{cross-attn-proj}

In the cross-attention-based approach, visual hidden states encoded by the visual encoder are used to condition a frozen LLM using freshly initialized cross-attention layers which are interleaved between the pretrained LLM layers \cite{laurençon2024mattersbuildingvisionlanguagemodels}. Unlike self-attention, cross-attention approach enables a separate consideration of prompt and visual document tokens. Flamingo \cite{flamingo2021} pioneered this approach with its Perceiver Resampler, which has since been adopted in various VrDU models (see Table~\ref{tab:v-only}). 

An advantage of using cross-attention is that it allows to process longer sequences from the encoder, and thus to use only high-resolution representations. For instance,  CogAgent~\cite{hong2024cogagentvisuallanguagemodel} employs a high-resolution encoder connected to the decoder through a cross-attention layer, while using self-attention with a low resolution version of the image.  %Similarly, Mini-Gemini \cite{mini-gemini} utilizes a Patch Info Mining module, where a low-resolution embedding provides a global view and acts as a query to extract relevant details from high-resolution patches, effectively incorporating fine-grained visual details.

In other words, cross-attention approaches for integrating visual features into LLMs enable the query/prompt tokens to explicitly interact with visual features, effectively leveraging the LLM’s capabilities. 

However, these methods require the introduction of many new parameters, as cross-attention layers are interleaved with the LLM’s architecture, significantly increasing the overall model size \cite{laurençon2024mattersbuildingvisionlanguagemodels}.

\subsection{Pretraining for visual features insertion}

\citet{mplugdocowl1.5} highlight that, to integrate visual features into an LLM, VrDU models must be pretrained on document parsing tasks. \citet{pix2struct, vary, nougat, mplugdocowl1.5, donut} exploit the fact that documents are often generated from a symbolic source document (e.g. HTML, latex, Markdown, extended Markdown format for table and charts or CSV/JSON) to convert document page screenshot into structured text for pretraining. \citet{mplugdocowl1.5} implements a multi-format reconstruction task named Unified Structured Learning.

\begin{table}[!t]
\centering
\tiny
\resizebox{\columnwidth}{!}{
   \begin{tabular}{|l|cc|cc|}
    \toprule
        \multirow{2}{*}{\textbf{Models}} & \textbf{Doc} & \textbf{Info} & \multirow{2}{*}{\textbf{DUDE}} & \textbf{MPDoc} \\
        & \textbf{VQA} & \textbf{VQA} & & \textbf{VQA} \\
        \midrule
        \rowcolor{LightCyan} \multicolumn{5}{c}{\textbf{T+L+V models (Section~\ref{sec:t+l+v})}} \\
        \midrule
        LayoutLMv3~\citeyear{layoutlmv3} & 83.4 & 45.1 & 20.3² & 55.3² \\
        ERNIE-Layout~\citeyear{ernie} & 88.4 & & & \\
        DocFormerv2~\citeyear{docformerv2} & 87.8 & 48.8 & 50.8² & 76.8² \\
        HiVT5~\citeyear{hivt5} & & & 23.0 & 62.0 \\
        GRAM~\citeyear{gram} & 86.0 & & 53.4 & \textbf{80.3} \\
        LayoutLLM~\citeyear{layoutllm} & 86.9 & & & \\
        DocLayLLM~\citeyear{doclayllm} & 78.4 & 40.9 & & \\
        TILT~\citeyear{tilt2021} & 87.1 & & & \\
        UDOP~\citeyear{udop} & 84.7 & 47.4 & & \\
        ViTLP~\citeyear{vitlp} & 65.9 & 28.7 & & \\
        Arctic-TILT~\citeyear{arctictilt} & \textbf{90.2} & \textbf{57.0} & \textbf{58.1} & \textbf{81.2} \\
        \midrule
        \rowcolor{LightCyan} \multicolumn{5}{c}{\textbf{Vision-only models (Section~\ref{sec:v-only})}} \\
        \midrule
        DONUT~\citeyear{donut} & 72.1 & 11.6 & & \\
        DESSURT~\citeyear{dessurt} & 63.2 & & & \\
        Pix2Struct~\citeyear{pix2struct} & 76.6 & 40.0 & & 62.0* \\
        SeRum~\citeyear{cao2023attentionmattersrethinkingvisual} & 77.9 & & & \\
        Kosmos2.5~\citeyear{kosmos25Paper} & 81.1 & 41.3 & & \\
        LLaVAR~\citeyear{llavar} & 6.73 & 12.3 & & \\
        Unidoc~\citeyear{unidoc} & 7.70 & 14.7 & & \\
        DocPedia~\citeyear{feng2024docpediaunleashingpowerlarge} & 47.8 & 15.2 & & \\
        CogAgent~\citeyear{hong2024cogagentvisuallanguagemodel} & 81.6 & 44.5 & & \\
        Vary~\citeyear{vary} & 76.3 & & & \\
        mPLUGDoc~\citeyear{ye2023mplugdocowlmodularizedmultimodallarge} & 62.2 & 38.2 & & \\
        QwenVL~\citeyear{bai2023qwenvlversatilevisionlanguagemodel} & 65.1 & 35.4 & & \textbf{84.4*} \\
        UREADER~\citeyear{ureader} & 65.4 & 42.2 & & \\
        Monkey~\citeyear{monkey} & 66.5 & 36.1 & & \\
        TextSquare~\citeyear{textsquare} & 84.3 & 51.5 & & \\
        TextMonkey~\citeyear{textmonkey} & 73.0 & 28.6 & & \\
        mPLUGDoc1.5~\citeyear{mplugdocowl1.5} & 82.2 & 50.7 & & \\
        ILMXC24KHD~\citeyear{internlmxcomposer24khd} & \textbf{90.0} & \textbf{68.6} & \textbf{56.1*} & 76.9* \\
        Idefics2~\citeyear{laurençon2024mattersbuildingvisionlanguagemodels} & 74.0 & & & 56.0* \\
        TextHawk~\citeyear{texthawk} & 76.4 & 50.6 & & \\
        TokenPacker~\citeyear{tokenpacker} & 70.0 & & & \\
        mPLUGDoc2~\citeyear{mplugdocowl2} & 80.7 & 46.4 & 46.8 & 69.4 \\
        HRVDA~\citeyear{hrvda} & 72.1 & 43.5 & & \\
        DocKylin~\citeyear{dockylin} & 77.3 & 46.6 & & \\
        \midrule
        \rowcolor{LightCyan} \multicolumn{5}{c}{\textbf{Commercial Models}} \\ 
        \midrule
        GPT-4V & 88.4 & \textbf{75.1} & & \\
        GPT-4o & \textbf{92.8} & & \textbf{54.0} & 67.0 \\
    \bottomrule
    \end{tabular}
}
\caption{Average Normalized Levenshtein Similarity (ANLS) on single and multi-page VQA. ² denotes Single-page-native models concatenating page representations for multi-page; * denotes models using a retriever (PDF-Wukong \cite{pdfwukong} for InternLMXComposer2-4KHD, \citet{multipageretrieval} for Pix2Struct, M3DocRAG \cite{cho2024m3docragmultimodalretrievalneed} for QwenVL and Idefics2). The top-3 scores are in bold.}
\label{tab:vqa-performance}
\end{table}

\section{Conclusion and Discussion}

While vision-only methods (Section~\ref{sec:v-only}) are gaining prominence in recent literature, they face significant challenges in balancing coarse and fine-grained VrD representations. This often results in excessive computational complexity or compression issues, making these methods unsuitable for multi-page document processing without a retriever (see Table~\ref{tab:vqa-performance}). For multi-page understanding, we argue that multi-modal approaches—combining textual, visual, and positional features—are more efficient (see Table~\ref{tab:vqa-performance}). 

In addition to the computational cost aspect, our view is that the community should prioritize developing methods to handle text, layout, and visual elements in documents, as we observe that documents are increasingly becoming digital-native, with bounding boxes and text readily accessible. However, these approaches remain challenging due to the need for effective alignment across textual and visual features, and due to the need for LLM to handle 2D positional information efficiently. 

To reduce redundant information between textual and visual features \cite{udop} and handle both information in an optimal way, we suggest focusing on integrating textual features within the visual representation using cross-attention mechanisms \cite{selfdoc} with text guiding the integration (query) when visual elements are less prominent in the document \cite{arctictilt}, and visual features guiding when visual elements are major in documents.

Our view is that the community should focus on developing methods to effectively process 2D information, exploring aspects such as granularity, the semantic connection to 2D positions, and multi-level attention mechanisms—both between semantically meaningful blocks and within those blocks, and adapting 2D position encoding to recent approaches \cite{su2023roformerenhancedtransformerrotary}. As shown in Table~\ref{tab:vqa-performance}, models that make extensive use of positional features—such as ERNIE-Layout \cite{ernie} and Arctic-TILT \cite{arctictilt}~--~have the best results. This indicates that text and layout information are essential for answering questions, even in complex charts and figures, making efficient layout handling critical.

\section{Limitations}

A first limitation of our survey lies in the lack of consistent evaluation across different techniques. While we discuss a range of methods—such as 2D position encoding strategies, approaches for integrating visual and textual information, projectors between the visual encoder output and the LLM decoder, sparse attention  approaches for multi-page document handling, ... -- these techniques are evaluated in their original experimental setups, which differ in terms of model architecture, training protocols, and datasets. As a result, it is challenging to draw definitive conclusions about which technique performs best in a given scenario. Although a fairer and more scientifically rigorous comparison would require benchmarking all methods under the same conditions, this was beyond the scope of our survey due to time and resource limitations.

A further limitation of this survey is that most of the comparisons in this survey are based on benchmarks for visual question answering (VQA), while we overlook several traditional document understanding tasks. These tasks include key information extraction, document layout analysis, document classification, or reading order prediction (beyond others), which are essential for many real-world applications such as automatic form processing, contract analysis, and archival document digitization. Our focus on VQA benchmarks is primarily motivated by their widespread use in recent research as a comprehensive testbed for evaluating VrDU approaches both in their information extraction and reasoning capabilities.

Additionally, we focus exclusively on transformer-based approaches. While this choice aligns with the current state of the art, it inevitably excludes earlier yet significant contributions. For instance, traditional methods leveraging LSTMs or Gated Recurrent Units have been  widely used in VrDU. More recent work has also started exploring alternative architectures such as state space models \cite{docmamba}. Graph-Based Relationship Modeling approaches, representing documents as hierarchical structures and employing graph neural networks (GNNs) to model relationships between document elements, are also extensively adopted by the community \cite{Dai2024GraphMLLMAG, Zhang2022MultimodalPB, Li2023EnhancingVD}. Due to space and scope constraints, we  focused on transformers, which dominate current research and offer a unified framework for integrating visual and textual modalities.

Finally, this survey focuses primarily on generic multi-element documents, such as PDFs and PowerPoint slides, as illustrated in Figure~\ref{fig:data_illustration} in appendix, rather than specific document types (e.g., tables, charts, or diagrams). Our decision to concentrate on general-purpose documents stems from the desire to provide a broad overview that covers documents combining multiple data types rather than diving into domain-specific challenges. Each specific domain—such as table understanding or chart interpretation—presents its own unique challenges and innovations, like cell, row and columns understanding for table, with approaches modeling column-wise and row-wise self-attention \cite{yin2020tabertpretrainingjointunderstanding, deng2020turltableunderstandingrepresentation}, derendering tasks for Charts, with approach converting chart image into their Matplotlib code \cite{alshetairy2024transformersutilizationchartunderstanding} with their associated JSON/CSV \cite{deplot}, or structure analysis tasks for diagrams, aiming at linking the legend to the diagram content \cite{tableandfigures}, %as well as subfigure separation tasks for segmenting subigure image panels automatically \cite{taschwer2016automaticseparationcompoundfigures}
which are beyond the scope of this survey.

\bibliography{main}

\appendix

\section{Example Appendix}
\label{sec:appendix}

\subsection{Visual Question Answering datasets}

We mainly focused on the Visual Question Answering (VQA) task in this survey as a benchmark to compare different models. document-VQA consists of answering a question based on the content of a document, requiring the model to understand both visual and textual information to provide an accurate response.

\begin{table*}[ht]
\tiny
\begin{tabular}{|l|ccccc|ccccc|} 
\hline
\multicolumn{1}{|c|}{} & \multicolumn{5}{c|}{\textbf{Document Characteristics (per document)}} & \multicolumn{5}{c|}{\textbf{Questions Characteristics}} \\
\multicolumn{1}{|c|}{\textbf{Datasets}} & \textbf{type} & \textbf{\#Pages} & \textbf{\#Tokens} & \textbf{\#Tab} & \textbf{\#Fig} & \textbf{Crosspage} & \textbf{Unans.} & \textbf{Crossdoc} & \textbf{\#Regions} & \textbf{Ans. length} \\ 
\hline
VisualMRC \citeyear{visualmrc} & Wikipedia pages & 1.0 & 151.46 & ? & ? & \ding{55} & \ding{55} & \ding{55} & \ding{55} & 9.55 \\ 
\underline{DocVQA} \citeyear{docvqa} & \href{https://www.industrydocuments.ucsf.edu/}{Industry Documents} & 1.0 & 182.8 & ? & ? & \ding{55} & \ding{55} & \ding{55} & \ding{55} & 2.43 \\
\underline{InfographicVQA} \citeyear{infographicvqa} & Posters (Canva, ...) & 1.2 & 217.9 & ? & ? & \ding{55} & \ding{55} & \ding{55} & \ding{55} & 1.6 \\
TAT-DQA \citeyear{tat-dqa} & \href{https://www.annualreports.com/}{Annual Reports} & 1.3 & 550.3 & >1 & ? & \ding{55} & \ding{55} & \ding{55} & \ding{55} & 3.44 \\
\underline{MP-DocVQA} \citeyear{hivt5} & \href{https://www.industrydocuments.ucsf.edu/}{Industry Documents} & 8.3 & 2026.6  & ? & ? & \ding{55} & \ding{55} & \ding{55} & \ding{55} & 2.2 \\
\underline{DUDE} \citeyear{dude} & \href{https://archive.org/}{archives}, \href{https://commons.wikimedia.org/}{wikimedia} & 5.7 & 1831.5  & ? & ? & \checkmark(2.1\%) & \checkmark(12.7\%) & \ding{55} & \ding{55} & 3.4 \\
SlideVQA \citeyear{slidevqa} & Slides from \href{https://www.slideshare.net/}{Slideshare} & 20.0 & 2030.5  & ? & ? & \checkmark(13.9\%) & \ding{55} & \ding{55} & \ding{55} & $\approx$1 \\
MMLongBenchDoc \citeyear{mmlongbenchdoc}& ArXiv, Reports, Tuto & 47.5 & 21214.1  & 25.4\% & 20.7\% & \checkmark(33.0\%) & \checkmark(22.5\%) & \ding{55} & \ding{55} & 2.8 \\
M3DocVQA \citeyear{cho2024m3docragmultimodalretrievalneed} & Wikipedia pages & 12.2 & ?  & ? & ? & \checkmark & ? & \checkmark(2.4k) & \ding{55} & ? \\
M-LongDoc \citeyear{Chia2024MLongdocAB} &  \href{https://www.manualslib.com/}{Manuals}, \href{https://www.annualreports.com/}{Reports} & 210.8 & 120988 & 71.8 & 161.1 & \ding{55} & \ding{55} & \ding{55} & \ding{55} & 180.3 \\
MMDocBench~\citeyear{zhu2024mmdocbenchbenchmarkinglargevisionlanguage} & Multi & 1.0 & ? & ? & ? & \ding{55} & \ding{55} & \ding{55} & 2.61 & 4.1 \\
BoundingDocs~\citeyear{giovannini2025boundingdocsunifieddatasetdocument} & Multi & 237k & ? & ? & ? & \checkmark & \ding{55} & \ding{55} & >=1 & >=1 \\ 
\hline
\end{tabular}
\caption{Overview of open-source Question-Answering VrDU datasets on PDFs or PPTs documents, summarizing document characteristics (e.g., average pages, tokens, tabs, figures per document) and question characteristics (e.g., presence of questions requiring cross-pages or cross-documents information, unanswerable questions, and average answer length). \#Region refers to the number of regions identified for answering questions in datasets with coordinate annotations. Underlined datasets are standard benchmarks used for model comparison in Table~\ref{tab:vqa-performance}.}
\label{tab:datasets}
\end{table*}

\begin{figure*}[ht]
    \centering
    \includegraphics[width=0.8\paperwidth, height=0.8\paperheight, keepaspectratio]{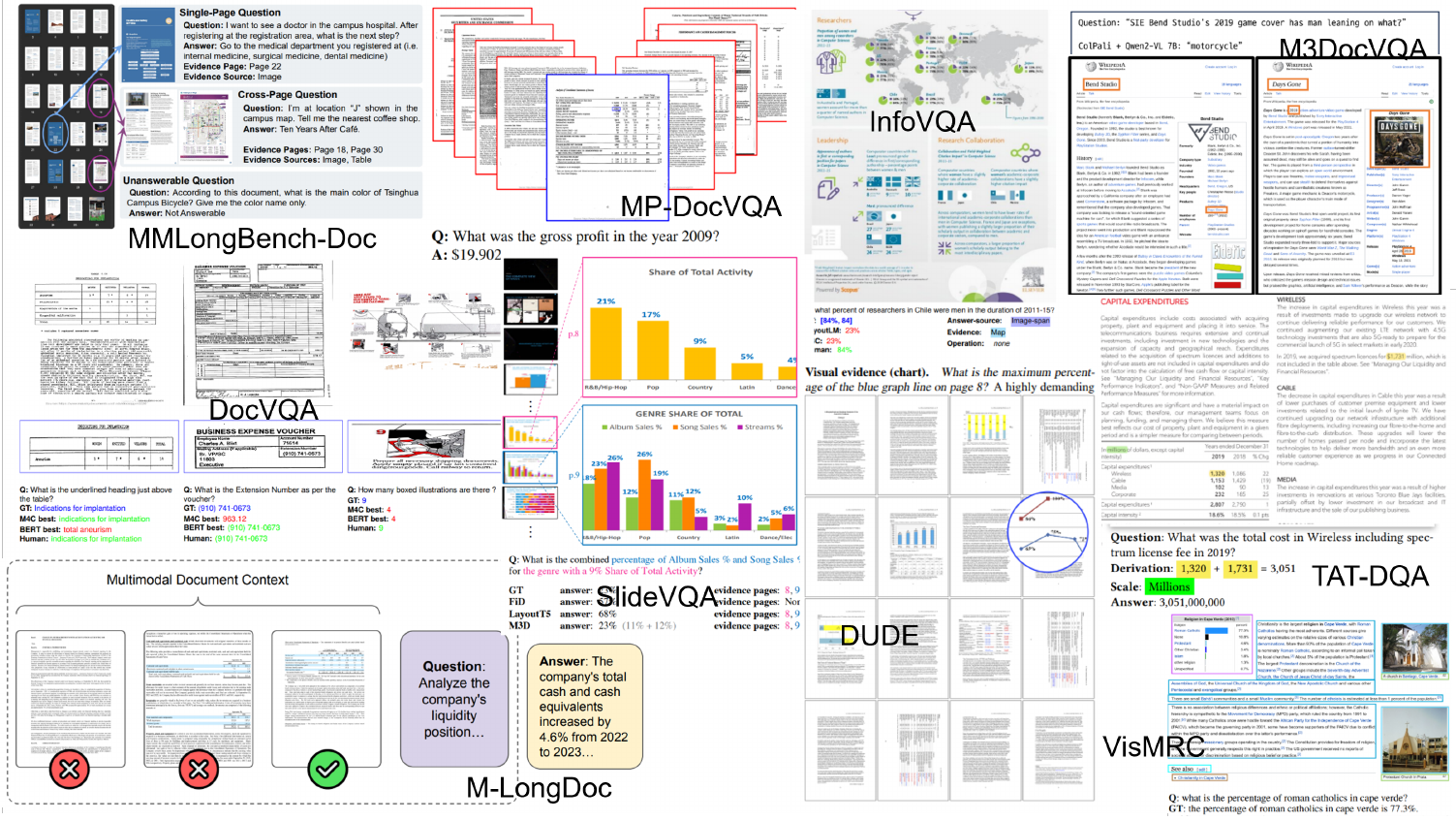}
    \caption{Illustration of the datasets listed in this survey}
    \label{fig:data_illustration}
\end{figure*}

\end{document}